%% file: main.tex
\documentclass[10pt,twocolumn,letterpaper]{article}

\usepackage[pagenumbers]{config/iccv} %

\input{config/preamble}

\definecolor{iccvblue}{rgb}{0.21,0.49,0.74}
\usepackage[pagebackref,breaklinks,colorlinks,allcolors=iccvblue]{hyperref}

\input{config/definitions}
\input{config/authors}

\title{\outTitleFULL}

\begin{document}

\twocolumn[{
\renewcommand\twocolumn[1][]{#1}
\vspace{-1.4 em}
\maketitle
\input{fig/01_teaser}
}]

\input{sec/00_Abstract}
\input{sec/01_Intro}
\input{sec/02_Related}
\input{sec/03_Method}
\input{sec/04_Results}

\input{sec/05_Conclusion}
\input{sec/06_Ack_Discl}

{
    \small
    \bibliographystyle{config/ieeenat_fullname}
    \bibliography{config/BIB}
}

\renewcommand{\thefigure}{S.\arabic{figure}}
\renewcommand{\thetable}{S.\arabic{table}}
\renewcommand{\theequation}{S.\arabic{equation}}
\setcounter{figure}{0}
\setcounter{table}{0}
\setcounter{equation}{0}

\input{sec/XX_SupMat}

\end{document}

%% file: config/preamble.tex
\usepackage[dvipsnames]{xcolor}
\newcommand{\red}[1]{{\textcolor{red}{#1}}}

\newcommand{\supmat}{\textcolor{black}{{Sup.~Mat.}}\xspace}

\newcommand{\nameCOLOR}[1]{\textcolor{black}{#1}}
\newcommand{\nameMethod}{\mbox{\nameCOLOR{SDFit}}\xspace}

\newcommand{\myEnum}[1]{\textcolor{black}{#1}}

\newcommand{\termCOLOR}[1]{\textcolor{black}{#1}}

\newcommand{\SDF}{\mbox{\termCOLOR{SDF}}\xspace}
\newcommand{\mSDF}{\mbox{\nameCOLOR{mSDF}}\xspace}
\newcommand{\mSDFs}{\mbox{\mSDF{s}}\xspace}

\newcommand{\DIT}{\mbox{{DIT}}\xspace}
\newcommand{\OpenShape}{\mbox{\termCOLOR{OpenShape}}\xspace}
\newcommand{\ZeroShape}{\mbox{\termCOLOR{ZeroShape}}\xspace}
\newcommand{\SDFusion}{\mbox{\termCOLOR{SDFusion}}\xspace}
\newcommand{\OpenPose}{\mbox{\termCOLOR{OpenPose}}\xspace}

\newcommand{\NA}{\mbox{\textcolor{gray}{N/A}}\xspace}

\newcommand{\outTitleFULL}{\vspace{-0.25 em}SDFit: 3D Object Pose and Shape by Fitting a Morphable SDF to a Single Image\vspace{-0.25 em}}

\renewcommand{\etal}{\mbox{et al.}\xspace}
\renewcommand{\ie}{\mbox{i.e.}\xspace}
\renewcommand{\eg}{\mbox{e.g.}\xspace}
\renewcommand{\wrt}{\mbox{w.r.t.}\xspace}

\newcommand{\zheading}[1]{\textbf{#1:}}
\newcommand{\qheading}[1]{\noindent\textbf{#1:}}

\newcommand{\smpl}{\mbox{\termCOLOR{SMPL}}\xspace}

\newcommand{\smplify}{\mbox{\termCOLOR{SMPLify}}\xspace}

\newcommand{\hps}{\mbox{HPS}\xspace}
\newcommand{\HPS}{\hps}
\newcommand{\ops}{\mbox{OPS}\xspace}
\newcommand{\OPS}{\ops}

\newcommand{\groundtruth}{{ground-truth}\xspace}

\newcommand{\stateoftheart}{{state-of-the-art}\xspace}
\newcommand{\sota}[0]{\mbox{SotA}\xspace}

\newcommand{\inthewild}{{in-the-wild}\xspace}

\usepackage{tikz}
\usepackage{graphicx}
\usepackage{overpic}
\usepackage{multirow}

\usepackage{pifont}
\newcommand{\cmark}{\color{ForestGreen}\ding{51}}

\newcommand{\xmark}{\color{red}\ding{55}}

\usepackage{fontawesome}

\usepackage{colortbl}

\usepackage{balance}

%% file: config/definitions.tex
\usepackage{amsmath}
\usepackage{amssymb}
\usepackage{xspace}
\usepackage{enumitem}
\DeclareMathOperator*{\argmax}{arg\,max}
\DeclareMathOperator*{\argmin}{arg\,min}

\definecolor{cvprblue}{RGB}{0, 113, 188}

\newcommand{\shapefunc}[0]{f^{sdf}_{\theta}}

\newcommand{\lookupfunc}[0]{f^{db}}

\newcommand{\diffusionfeat}[0]{\cF^{diff}}
\newcommand{\imgfeat}[0]{\cF_{\img}}

\newcommand{\dinofeat}[0]{\cF^{DINOv2}\xspace}

\newcommand{\gtmask}[0]{\cM\xspace}
\newcommand{\gtdepth}[0]{\cD\xspace}
\newcommand{\gtnormals}[0]{\cN\xspace}

\newcommand{\img}[0]{\cI\xspace}
\newcommand{\maskedimgfeat}[0]{\mathbf{\cF}_{M(\img)}\xspace}

\newcommand{\teximg}[0]{\img^{tex}\xspace}

\newcommand{\warpernet}[0]{W\xspace}
\newcommand{\templatenet}[0]{T\xspace}

\newcommand{\similaritymatrix}[0]{\mathbf{\cA}\xspace}

\newcommand{\lookupmethod}[0]{\mbox{\termCOLOR{OpenShape}}\xspace}
\newcommand{\shapenet}[0]{\mbox{ShapeNet}\xspace}
\newcommand{\ShapeNet}[0]{\shapenet}

\newcommand{\controlnet}[0]{\mbox{ControlNet}\xspace}
\newcommand{\stablediffusion}[0]{\mbox{StableDiffusion}\xspace}
\newcommand{\sd}[0]{\mbox{\termCOLOR{SD v1.5}}\xspace}
\newcommand{\dino}[0]{\mbox{\termCOLOR{DINOv2}}\xspace}
\newcommand{\difff}[0]{\mbox{\termCOLOR{Diff3F}}\xspace}

\newcommand{\sdfusion}[0]{\mbox{\termCOLOR{SDFusion}}\xspace}
\newcommand{\zeroshape}[0]{\mbox{\termCOLOR{ZeroShape}}\xspace}
\newcommand{\triposr}[0]{\mbox{TripoSR}\xspace}

\newcommand{\classzeroshape}[0]{\mbox{\zeroshape-{\tt CLS}}\xspace}

\newcommand{\flexicubes}[0]{\mbox{\termCOLOR{FlexiCubes}}\xspace}

\newcommand{\pixTD}{\mbox{{Pix3D}}\xspace}
\newcommand{\comic}{\mbox{{COMIC}}\xspace}
\newcommand{\pascalTD}{\mbox{{Pascal3D+}}\xspace}

\newcommand{\imgtoshape}[0]{pixel-vertex\xspace}

\newcommand{\griddim}[0]{N}
\newcommand{\latentdim}[0]{256}
\newcommand{\featdim}[0]{2368}

\newcommand{\shapecollection}[0]{\mathbf{\cS}}
\newcommand{\latentdb}[0]{\mathbf{\cZ}}

\newcommand{\initshape}[0]{S_{init}}
\newcommand{\initR}[0]{R_{init}}
\newcommand{\initT}[0]{t_{init}}
\newcommand{\initS}[0]{s_{init}}

\newcommand{\shapefeat}[0]{\mathbf{\cF}_{S}}

\newcommand{\lat}[0]{z}
\newcommand{\latinit}[0]{z_{init}}

\newcommand{\nviews}[0]{J}
\newcommand{\scale}{\mathbf{s}}

\newcommand{\niterpose}[0]{300}
\newcommand{\nitershape}[0]{1000}
\newcommand{\griddimnum}[0]{32}
\newcommand{\avgtime}[0]{$3$ min\xspace}

\newcommand{\predmask}[0]{\widehat{\cM}}
\newcommand{\predmaski}[0]{\predmask^i}
\newcommand{\preddepth}[0]{\widehat{\cD}}
\newcommand{\preddepthi}[0]{\preddepth^i}
\newcommand{\prednormals}[0]{\widehat{\cN}}
\newcommand{\prednormalsi}[0]{\prednormals^i}
\newcommand{\predcontour}[0]{\widehat{\cC}}
\newcommand{\predcontouri}[0]{\predcontour^i}
\newcommand{\gtcontour}[0]{\cC}

\usepackage{amsthm}

\newcommand{\colorRef}[1]{\textcolor{black}{#1}}
\usepackage[capitalize]{cleveref}
\crefname{figure}{\colorRef{Fig.}}{\colorRef{Figs.}}
\Crefname{figure}{\colorRef{Figure}}{\colorRef{Figures}}
\crefname{section}{\colorRef{Sec.}}{\colorRef{Secs.}}
\Crefname{section}{\colorRef{Section}}{\colorRef{Sections}}
\Crefname{table}{\colorRef{Table}}{\colorRef{Tables}}
\crefname{table}{\colorRef{Tab.}}{\colorRef{Tabs.}}
\Crefname{equation}{\colorRef{Equation}}{\colorRef{Equations}}
\crefname{equation}{\colorRef{Eq.}}{\colorRef{Eqs.}}

\usepackage[normalem]{ulem}

\newcommand{\nR}{\mathbb{R}}

\newcommand{\cA}{\mathcal{A}}

\newcommand{\cC}{\mathcal{C}}
\newcommand{\cD}{\mathcal{D}}

\newcommand{\cF}{\mathcal{F}}

\newcommand{\cI}{\mathcal{I}}

\newcommand{\cM}{\mathcal{M}}
\newcommand{\cN}{\mathcal{N}}

\newcommand{\cS}{\mathcal{S}}

\newcommand{\cZ}{\mathcal{Z}}

\DeclareOldFontCommand{\bf}{\normalfont\bfseries}{\mathbf}

\newcommand{\FScore}{{F-Score}\xspace}

\DeclareSymbolFont{matha}{OML}{txmi}{m}{it}
\DeclareMathSymbol{\varv}{\mathord}{matha}{118}

%% file: config/authors.tex
\author{
Dimitrije Anti\'{c}$^{1}$   \quad 
Georgios Paschalidis$^{1}$  \quad 
Shashank Tripathi$^{2}$     \\
Theo Gevers$^{1}$           \quad 
Sai Kumar Dwivedi$^{2}$     \quad 
Dimitrios Tzionas$^1$       \\
{\small
$^1$University of Amsterdam, The Netherlands \quad
$^2$Max Planck Institute for Intelligent Systems, T{\"u}bingen, Germany
}\\
{\tt\small \{d.antic,g.paschalidis,th.gevers,d.tzionas\}@uva.nl}
\quad 
{\tt\small \{sdwivedi,stripathi\}@tue.mpg.de}
}

%% file: fig/01_teaser.tex
\begin{center}
    \vspace{-0.8 em}
    \centering
    \captionsetup{type=figure}
        \includegraphics{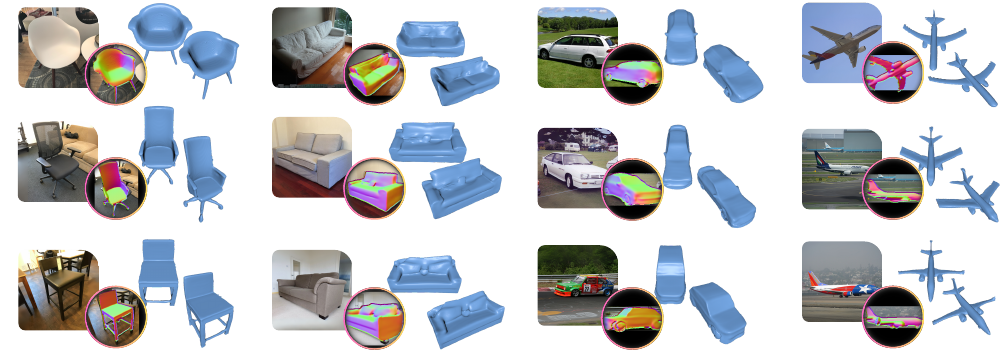}
    \vspace{-0.5 em}
    \caption{
                We present \nameMethod, a novel framework that recovers an object's 3D pose and shape from a single image. 
                To this end, \nameMethod uses a learned, category-level, morphable \SDF (\mSDF) shape model, 
                namely \DIT \cite{dit}, and fits this to images in a render-and-compare (a.k.a.~analysis-by-synthesis) fashion. \nameMethod is robust to occlusions and uncommon poses, and requires no retraining for in-the-wild images. 
                For visualizing the fitted \mSDF overlaid on the input image, we show the \mSDF's normals with color coding.
                \faSearch~\textbf{Zoom in} for details.
    }
    \label{fig:teaser}
    \vspace{+1.2 em}
\end{center}

%% file: sec/00_Abstract.tex
\begin{abstract}
Recovering 3D object pose and shape from a single image is a challenging and %
ill-posed problem. 
This is due to strong (self-)occlusions, depth ambiguities, the vast intra- and inter-class shape variance, and the lack of 3D ground truth for natural images. 
Existing deep-network methods are trained 
on synthetic datasets to predict 3D shapes, so they often struggle generalizing to real-world images. 
Moreover, they lack an explicit feedback loop for refining noisy estimates, and primarily focus on geometry without directly considering pixel alignment. 
To tackle these limitations, 
we develop a novel render-and-compare optimization framework, called \nameMethod. 
This has three key innovations: 
First, it uses a learned category-specific and morphable signed-distance-function (\mSDF) model, 
and fits this to an image by iteratively refining both 3D pose and shape. 
The \mSDF robustifies inference by constraining the search on the manifold of valid shapes, while allowing for arbitrary shape topologies. 
Second, \nameMethod retrieves an initial 3D shape that likely matches the image, by exploiting foundational models for efficient look-up into 3D shape databases. 
Third, \nameMethod initializes pose by establishing rich 2D-3D correspondences between the image and the \mSDF through foundational features.
We evaluate \nameMethod on three image datasets, \ie, \pixTD, \pascalTD, and \comic. 
\nameMethod performs on par with \sota feed-forward networks for unoccluded images and common poses, but is uniquely robust to occlusions and uncommon poses. 
Moreover, it requires no retraining for unseen images.
Thus, \nameMethod contributes new insights for generalizing in the wild. 
Code is available at \url{https://anticdimi.github.io/sdfit}.
\end{abstract}

%% file: sec/01_Intro.tex
\section{Introduction}
\label{sec:intro}

Recovering 3D object pose and shape (\OPS) from single images is key for building intelligent systems and mixed realities. 
However, the task is highly ill-posed due to strong challenges such as depth ambiguities, \mbox{(self-)}occlusions, and the huge variance in shape, appearance, and viewpoint. 
Yet, humans routinely solve this task by building and exploiting rich prior models through experience. 
Despite progress, computers still lack reliable methods and priors for reconstructing 3D objects from natural images. 
Our goal is to recover 3D object shape and pose from a natural image. 

\pagebreak

To this end, we draw inspiration from the ``analogous'' task of human pose and shape (\HPS) estimation. 
Morphable generative body models \cite{anguelov2005scape, SMPL-X:2019, xu2020ghum, Joo2018_adam} such as \smpl \cite{SMPL:2015} make \HPS relatively reliable.
Such models are data-driven and capture shape variance across a database of body scans. 
When fitting such models to single images \cite{SMPL-X:2019, xu2020ghum, Xiang_2019_CVPR}, \eg, through the \smplify~\cite{bogo2016simplify} method, they act as a \emph{strong shape prior}. 
That is, full-body shape can be reliably inferred even when bodies are \emph{partially occluded}.  
Such occlusions are also common for object images taken in the wild.

However, perhaps counter-intuitively, there exists no \smpl-like model or \smplify-like method for objects. 
But we cannot trivially adapt \HPS methods for solving \OPS as, despite commonalities, these tasks differ in three 
key ways: 
\myEnum{(1)}~\emph{Shape variance} is much bigger for objects (which is both intra- and inter-class) than for bodies (which is only intra-class).
For example, an armchair looks different from an airplane, but also from an office chair or a folding chair. 
\myEnum{(2)}~Objects have a wildly \emph{varying topology} (\eg, chairs with a varying number of legs) while bodies have the same one. 
\myEnum{(3)}~To guide \HPS fitting, \OpenPose-like methods \cite{cao2019openpose, Lugaresi2019MediaPipeAF} robustly detect in images 2D joints that directly correspond to 3D \smpl joints. 
In contrast, for general objects, detecting \emph{correspondences} between 2D images and a textureless 3D model (let alone a \emph{morphable} 3D model) is an open problem. 
Thus, \OPS and \HPS methods have evolved separately.

The current \OPS paradigm is rendering synthetic images from 3D databases \cite{omniobject3d, objaverse, objaverse_xl} for training deep networks to regress 3D shape from an image \cite{alwala2022ss3d, huang2023zeroshape, huand2023shapeclipper, hong2024lrm, TripoSR2024}, or to generate it via image-conditioned diffusion \mbox{\cite{cheng2023sdfusion, deng2023nerdi, kyriazy2023pc, kyriazi2023realfusion, shape, pointe}}. 
Such methods work well for in-distribution, unoccluded images, and common poses, but have three limitations:
\myEnum{(1)}~They~struggle generalizing to natural-looking, out-of-distribution images with occlusions and uncommon poses. 
\myEnum{(2)}~They mostly perform only feed-forward inference, and lack an explicit feedback loop for refining noisy estimates. 
\myEnum{(3)}~They mostly focus on geometry alone, largely ignoring object or camera pose, and by extension, pixel alignment.

Tackling the above limitations requires a strong shape prior for constraining the search to plausible shapes, \ie, for generating and refining plausible shape hypotheses.
To this end, we exploit a category-level morphable signed-distance function (\mSDF) model that generates 3D shape hypotheses through sampling its latent space (similar to \smpl \cite{SMPL:2015}); here we use \DIT~\cite{dit}. 
This encodes the manifold of valid shapes, while allowing arbitrary topologies \cite{deepsdf,dit,sitzmann2020siren} and establishing dense correspondences across morphed shapes. 

We exploit this to develop \nameMethod, a novel framework that fits the \mSDF to an image (like \smplify does for \smpl) by searching for a latent shape code and pose that best ``matches'' image cues; for an overview see \cref{fig:intro}.  
This has been done for 3D point clouds \cite{li2023ddit} but not for 2D images, which is much more challenging. We fill this gap here. 

However, fitting an \mSDF to an image is challenging not only due to depth ambiguities, but also due to requiring a good 3D shape and pose initialization, which is still unsolved. 
To~initialize shape, we exploit \OpenShape's~\cite{liu2023openshape} multimodal latent space to retrieve an \mSDF shape that matches the image; this is fast and scales to large databases \cite{objaverse}. 
To initialize pose, we decorate the initial shape with foundational features~\cite{rombach2021highresolution, Zhang_2023_ICCV, oquab2024dinov2}, and match these to features extracted from the image. 
This produces 2D-3D correspondences, used to recover pose.
The above has been done only for fixed shapes (3D meshes) \cite{cseke_tripathi_2025_pico, dwivedi_interactvlm_2025, ornek2024foundPose}, so they are novel for morphable shapes (\mSDFs). 
Eventually, our framework refines both 3D pose and shape via optimization with a feedback loop, \ie, it iteratively refines the \mSDF hypothesis to minimize the discrepancy between respective \mSDF-rendered and image-extracted feature maps, until convergence; see example reconstructions in \cref{fig:teaser}.

\input{fig/02_highlevel_overview}

We evaluate on three datasets \cite{pix3d,pascal3d+,chord} for 3D shape estimation (with and without occlusions), and for image alignment that involves both shape and pose estimation. 
Evaluation shows that our \nameMethod fitting framework performs on par with strong feed-forward regression-~\cite{huang2023zeroshape} and diffusion-based~\cite{cheng2023sdfusion,TripoSR2024} baselines for unoccluded images. 
However, \nameMethod excels under occlusions, while requiring no re-training for out-of-distribution images. 
Note that \nameMethod uniquely treats both pose and shape as first-class citizens.

In summary, the main contributions of our work are:
\myEnum{(1)}~A novel framework (\nameMethod) that uses a 3D morphable \SDF (\mSDF) model as a strong 3D shape prior, and fits this to a single image, while being uniquely robust to occlusions. 
\myEnum{(2)}~A novel \mSDF shape initialization, casted as a retrieval problem in a joint latent space of 2D images and 3D shapes. 
\myEnum{(3)}~A novel \mSDF pose initialization, using foundational models to establish rich image-to-\mSDF correspondences. 

Code for \nameMethod is available for research purposes at \url{https://anticdimi.github.io/sdfit}.

%% file: fig/02_highlevel_overview.tex
\begin{figure}
    \centering
        \includegraphics[width=0.99 \columnwidth]{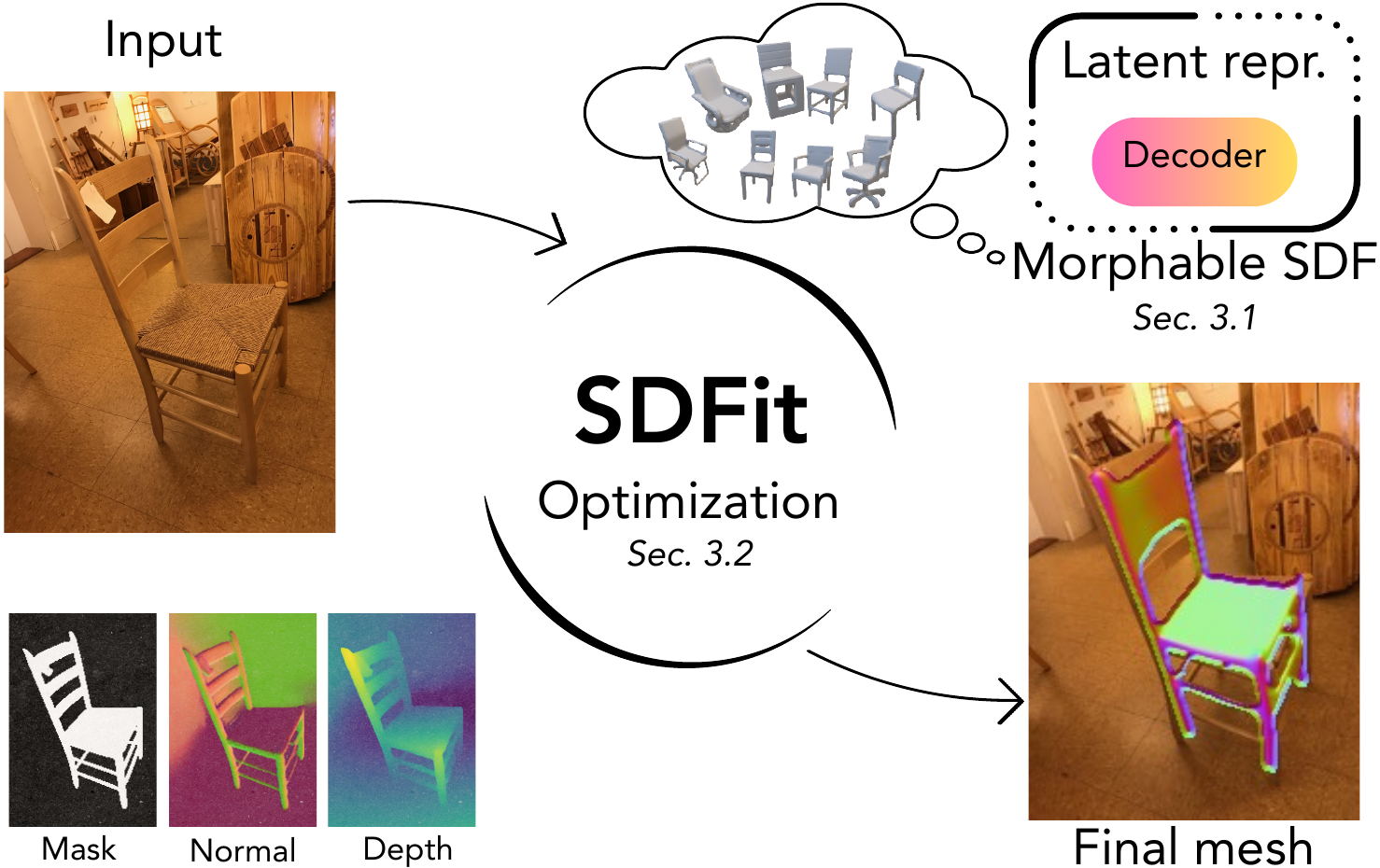}
        \vspace{-0.7 em}
        \caption{
            High-level 
            overview of our \nameMethod framework. 
            To recover both 3D object pose and shape, we fit a morphable signed-distance function (\mSDF) model to observed image features (\ie, extracted normal, depth and binary masks) in a render-and-compare fashion. 
        }
    \label{fig:intro}
    \vspace{-0.7 em}
\end{figure}

%% file: sec/02_Related.tex
\section{Related Work}

\hspace{\parindent}
\zheading{Object Shape Estimation}
Recent work on 3D shape inference from images represents shape in two main ways:
(1)~via explicit representations like voxel grids~\cite{rw_voxel1, rw_voxel2}, point clouds~\cite{rw_pc1, rw_pc2}, polygonal meshes~\cite{huand2023shapeclipper, alwala2022ss3d, meshrcnn, pixel2mesh} and 
(2)~via implicit representations like Neural Radiance Fields (NeRF)~\cite{rw_nerf1, nerf-image} or Signed-Distance Fields (SDF)~\cite{deepsdf, dit}. The former is easier to model but struggles with complex structures, while the latter provides more compact and flexible alternatives by encoding shapes as continuous fields.
We follow the latter, and specifically SDFs.

Approaches for 3D shape estimation follow three main paradigms, \ie, they are based on regression~\cite{alwala2022ss3d, huand2023shapeclipper, ye2021shelf, hong2024lrm, huang2023zeroshape, TripoSR2024}, generation~\cite{liu2023zero1to3, shape, pointe, kyriazi2023realfusion, cheng2023sdfusion} or retrieval~\cite{liu2023openshape, zhou2024uni3D}. 

Regression methods have significantly advanced 3D shape reconstruction from single images. 
This includes methods like SS3D~\cite{alwala2022ss3d}, which is pretrained on \mbox{ShapeNet}~\cite{chang2015shapenet} and fine-tuned on real-world images, leveraging category-level models for better performance. 
\mbox{ShapeClipper}~\cite{huand2023shapeclipper} enhances this with CLIP-based shape consistency.
Similarly, LRM~\cite{hong2024lrm} and TripoSR~\cite{TripoSR2024} predict NeRF using a transformer, achieving detailed 3D reconstruction. 
Recently, \mbox{ZeroShape}~\cite{huang2023zeroshape} infers camera intrinsics and depth as proxy states to improve reconstruction. 
However, these models often struggle generalizing to unseen categories and capturing the full diversity of complex or real-world shapes.

Generative methods, such as Zero123~\cite{liu2023zero1to3}, leverage foundational models for 3D shape estimation utilizing diffusion models to generate novel views from a single image, which are then used in multiview-to-3D methods such as \mbox{One-2-3-45}~\cite{one-2-3-45, magic123}. 
However, appearance quality (which is usually the priority) trades off against geometry quality. 
\mbox{SDFusion}~\cite{cheng2023sdfusion} learns an image-conditioned diffusion process on the latent representation of the object SDF. 

Retrieval methods, such as \mbox{OpenShape}~\cite{liu2023openshape}, align multimodal data, such as images and point clouds. 
Then, given an image, they retrieve the closest-looking 3D object from a database.
However, 3D databases have finite sizes, thus, retrieved shapes might not accurately match input images. 
Yet, this approach is fast and scales well to large databases, so we exploit this in our work for shape initialization. 

\zheading{Object Pose \& Shape Estimation}
Recent methods on single-image object pose estimation perform either direct pose parameter estimation~\cite{meshrcnn, rw_pose1} or alignment of a 3D template model with an input modality (\eg, image, features, keypoints)~\cite{rw_pose2, rw_pose3, rw_pose4, rw_pose5}. 
The former methods directly regress rotation, translation, and scale. 
The latter ones predict either sparse~\cite{rw_pose2, rw_pose3} or dense 3D-3D~\cite{rw_pose4} correspondences, or dense 2D-3D correspondences~\cite{rw_pose5,ornek2024foundPose}, and exploit these to solve for pose via the PnP~\cite{pnp} algorithm. 
While effective, this depends on accurate camera or depth data, while also requiring an a-priori known shape. 
We take this approach to initialize the pose of our initial shape. 

More recently, ROCA~\cite{gumeli2022roca} jointly estimates object pose and shape. 
To this end, it improves the pose estimate via differentiable Procrustes optimization on a retrieved CAD model.
However, the fixed shape of CAD models compromises reconstruction. 
Similarly, Pavllo~\etal~\cite{nerf-image} also estimate pose and shape using NeRFs, without any refinement. 
In contrast, \nameMethod optimizes both pose and shape using 3D-aware feature ``decoration'' through foundation models.

\zheading{3D-aware Foundational Models}
Large foundational models have catalized many 2D vision tasks~\cite{awais2023foundational}. 
Banani \etal~\cite{banani2024probing} find that \mbox{DINOv2}~\cite{oquab2024dinov2} and \mbox{StableDiffusion}~\cite{rombach2021highresolution} features also facilitate 3D tasks. 
We use features from these models to establish dense image-to-3D correspondences.

%% file: sec/03_Method.tex
\section{Method}
    We recover 3D object pose and shape from a single image via a novel render-and-compare framework, called \nameMethod; for an overview see \cref{fig:method}. 
    At the core of this lies a 3D morphable signed-distance function (\mSDF) model (\cref{sub:shape_prior}), and exploiting recent foundational models~\cite{oquab2024dinov2, Zhang_2023_ICCV, rombach2021highresolution, liu2023openshape}. 

    Our \nameMethod framework fits the \mSDF to image cues (\cref{sub:fitting}) by jointly optimizing over its shape and pose. 
    However, optimization-based methods are prone to local minima, so they need a good initialization. 
    To this end, \nameMethod first initializes the \mSDF shape through a \stateoftheart (\sota) retrieval-based technique (\cref{sub:shape_init}). 
    Then, it initializes pose by aligning the initial shape to rich, \sota foundational features extracted from the image (\cref{sub:pose_init}). 

    \input{fig/03_method}

\subsection{Shape Representation}
\label{sub:shape_prior}

    We represent 3D object shape via a learned, category-level, morphable signed-distance function (\mSDF) model. 
    
    \zheading{\mSDF} 
    Here we use the %
    \DIT model~\cite{dit}. 
    Each shape is encoded by a unique latent code, \mbox{$\lat \in \nR^{\latentdim}$}, in a compact space learned by auto-decoding a 3D dataset~\cite{chang2015shapenet}. 
    Mapping any 3D point, $x$, to a signed distance is parameterized by a network $\shapefunc: \nR^{3 \times \latentdim} \rightarrow \nR$ (with weights $\theta$) conditioned on \mbox{latent $\lat$}. 
    Each 3D shape, $S$, is encoded as the \mSDF's $0$-level set, $S = \{x \in \mathbb{R}^3 \mid \shapefunc(x; \lat) = 0\}$. 

    \DIT decodes a latent $\lat$ into signed distances through a warping function, $\warpernet(x; \lat)$, that ``warps" any 3D point, $x$, to a canonical space defined by a learned \SDF template, $\templatenet$. 
    This models the inter-category shape variance \wrt the template, and defines dense correspondences to it. 
    Note that training DIT comes with a useful byproduct, that is, it yields a collection of latent codes, $\latentdb$, for all training shapes $\lat$. 
    We use these later to initialize the shape hypothesis (\cref{sub:shape_init}). 
    
    \input{fig/04_method_canonloss}

    \zheading{Rendering}
    Rendering an \mSDF is not straightforward, so we extract a 3D mesh as a proxy that we exploit for differentiable rendering. 
    In each iteration we take three steps: 
    (1)~we predict SDF values via \mbox{$\shapefunc$ on a 3D grid}, 
    (2)~we extract a mesh using \mbox{FlexiCubes}~\cite{shen2023flexicubes}, and 
    (3)~we pose it by applying a \mbox{6-DoF} rigid transformation $(R, t) \in \text{SE}(3)$.

\subsection{Fitting Pose \& Shape}
\label{sub:fitting}
    To recover \OPS from an image, \nameMethod optimizes over object shape, $\lat \in \nR^{\latentdim}$, scale, $\scale \in \nR^3$, and pose, \mbox{$(R, t) \in \text{SE}(3)$}, by minimizing via render-and-compare the energy function:
    \begin{align}\label{eq:objectiveFULL}
        E =                      E_\gtmask + 
            \lambda_\gtnormals   E_\gtnormals + 
            \lambda_\gtdepth     E_\gtdepth + 
            \lambda_{DT}         E_{DT} + 
            \lambda_{R}          E_{R} \text{,}
    \end{align}
    where 
    $\gtmask$ is the mask, 
    $\gtnormals$ the normal map, 
    $\gtdepth$   the depth map, 
    DT denotes a 2D distance transform, 
    R denotes regularization, and 
    $\lambda$ are steering weights. 

    \noindent
    The individual energy terms are:    \begin{align}\label{eq:objectiveTERMS}
        E_\gtmask     =~ & \text{MSE}(\predmaski, \gtmask) + \lambda_{IoU} \cdot IoU(\predmaski, \gtmask),\\
        E_\gtdepth    =~ & \text{SSI-MAE}(\preddepthi, \gtdepth),    \label{eq:depth}\\
        E_\gtnormals  =~ & \text{MSE}(\prednormalsi, \gtnormals),\\
        E_{DT} =~ & \sum\nolimits_{\hat{x} \in \predcontouri} \min_{x \in \gtcontour} \|\hat{x} - x\|_1.
    \end{align}
    where 
    non-hat symbols are ``ground-truth'' observations, 
    hat denotes maps rendered from the running \mSDF hypothesis, 
    $i$ is the running iteration, 
    $\gtcontour$ the mask contour, 
    \mbox{MSE} the mean squared error, 
    \mbox{IoU} the intersection-over-union, while 
    \mbox{SSI-MAE} is a scale- and shift-invariant depth loss ~\cite{ranftl2022midas}.

    To regularize fitting under self-occlusions, %
    a regularization loss, $E_{R}$, encourages the running shape hypothesis, $S_i$, to be consistent with the initial estimate, $S_{init}$ (\cref{sub:shape_init}). 
    A simple way for this is to penalize deviation of the running $\lat$ code from the code $\lat_{init}$ of $S_{init}$, but, empirically, this causes local minima when $S_{init}$ has a wrong topology (\eg, a chair that erroneously misses armrests).

    Instead, \nameMethod geometrically regularizes to $S_{init}$ so it can still refine the topology (\eg, chairs growing missing armrests).  
    To this end, it uses the correspondences of $S$ and $S_{init}$ to the template, $\templatenet$, to map each vertex $x \in S_i$ to the closest vertex $u \in \initshape$. 
    Specifically, as shown in \cref{fig:method_canonloss},
    \mbox{(1)~it warps} $S_i$          vertices on the template in \emph{canonical space}, 
    \mbox{(2)~it warps} $\initshape$   vertices on the same template as well, and 
    \mbox{(3)~for each} warped vertex of $S_i$ it finds the closest warped vertex of $\initshape$, and eventually 
    \mbox{(4)~computes the MSE} for corresponding vertices in \emph{world space}. 
    In technical terms: 
    \begin{align}
        E_{R} &= MSE(S_i, S_{init}) \label{eq:regularization}                      \text{,}  
        \\
        S_{init} &= \{ \varv \mid 
                        \argmin_{\varv \in S_{init}} \|W(\varv; \lat_{init}) - W(x; \lat_{i})\|_2\}      \text{,}
        \label{eq:regLoss}
    \end{align}
    where 
    $x \in S_i$                 are vertices of $S_i$, 
    $W(x; \lat_{i})$            are these vertices mapped into the canonical space via the warper
    $W(\cdot)$, 
    $\varv \in S_{init}$        are vertices of $S_{init}$, 
    $W(\varv; \lat_{init})$     are these vertices mapped into the canonical space, and 
    $S_i=\shapefunc(x; \lat_i)$  is the \mbox{$i$-th}  iteration shape hypothesis (in canonical space). 
    
    During optimization, in each iteration, \nameMethod evaluates the energy function $E$ of \cref{eq:objectiveFULL}, backpropagates gradients, and updates the hypothesis parameters $\lat_i$, $R_i$, and $t_i$.

\subsection{Shape Initialization}
\label{sub:shape_init}

    \nameMethod initializes 
    the shape code, $\lat$, by exploiting the retrieval-based \lookupmethod~\cite{liu2023openshape} model, denoted as $\lookupfunc$. 
    This encodes multiple modalities (images, 3D point clouds) into a joint latent space, and facilitates searching for the 3D object, $\initshape$, that best resembles an input image, $\img$, by: 
    (1)~embedding the shapes $\shapecollection$ of \mSDF training data via $\lookupfunc$; 
    (2)~embedding image $\img$ into the same latent space via $\lookupfunc$; 
    (3)~retrieving the shape whose embedding most closely lies to the image embedding. 
    More formally, the initial-shape latent code, $\lat_{init}$, is the code $\lat$ whose 3D shape embedding, $\lookupfunc(S_\lat)$ lies closest to the image embedding, 
    $\lookupfunc(\img)$, via the cosine-similarity metric:
    \begin{align}
        \latinit &= \argmax_{\lat \in \latentdb} \frac{\lookupfunc(\img) 
        \;\,
        \cdot 
        \;\,
        \lookupfunc(S_\lat)}{\|\lookupfunc(\img)\|_2 
        \;\,
        \|\lookupfunc(S_\lat)\|_2}
        \text{,}
    \end{align}
    where 
    $S_z$ is $0$-level set of $\shapefunc(x; \lat)$, 
    $\lat$ is the shape latent code, and 
    $\latentdb$ is a database of auto-decoded latent codes, each corresponding to a shape instance in the mSDF training set.

\subsection{Pose Initialization}
\label{sub:pose_init}

    To initialize 3D pose from a single view, \nameMethod: 
    (1)~establishes correspondences between 2D pixels and 3D points, 
    (2)~estimates camera intrinsics, 
    (3)~filters out noisy correspondences with RANSAC, and 
    (4)~applies the PnP method. 

    To find correspondences, \nameMethod computes image features from the input image and rendered mSDF images. 
    To this end, inspired from image-to-image matching, it leverages features from foundational models such as \stablediffusion (SD) \cite{rombach2021highresolution} (or \controlnet\cite{Zhang_2023_ICCV}) and \dino~\cite{oquab2024dinov2}. 
    Specifically, it computes hybrid features that combine \sd (\controlnet) and \dino ones, as these encode geometry and semantic cues~\cite{NEURIPS2023_tale2features,dutt2023diffusion,luo2023dhf} that are crucial for 3D understanding. 
    In detail, it establishes 2D-3D \imgtoshape correspondences as described in the following paragraphs.
    
    \zheading{Image Features}
    \label{subsub:img_features} 
        \nameMethod uses the pretrained \controlnet \cite{Zhang_2023_ICCV} and \dino \cite{oquab2024dinov2} models. 
        It conditions \controlnet on the prompt {\tt``\tt A <category>, photorealistic, real-world''}, as well as on normal and depth maps estimated from an image for inpainting \cite{dutt2023diffusion}, \ie, hallucinating the original image from condition signals. 
        Crucially, this pushes \controlnet to semantically differentiate between nearby pixels \cite{dutt2023diffusion}, so features extracted from its layers capture \emph{semantic} cues. 
        Then, it applies \dino\cite{oquab2024dinov2} on an image to extract features capturing \emph{geometric} cues \cite{banani2024probing}. 
        Last, it forms hybrid features by concatenating per pixel the complementary 
        \controlnet and \dino features. 

        In technical terms, for an input image $\img$, estimated \cite{kar20223d} normal and depth maps, $\gtnormals$ and $\gtdepth$, and a text prompt, \nameMethod uses a pretrained \controlnet to generate (inpaint) a ``textured'' image, $\teximg$. 
        To get \controlnet features, at the last diffusion step \nameMethod extracts features $\diffusionfeat_2$ and $\diffusionfeat_4$ from its \mbox{UNet}-decoder layers 2 and 4, respectively, upsamples these to the resolution of $\img$, and concatenates these to obtain the feature $\diffusionfeat = \{\diffusionfeat_2 || \diffusionfeat_4\}$. 
        Note that here features from early layers emphasize semantic and geometric cues over texture ones ~\cite{dutt2023diffusion,banani2024probing}, which is beneficial as our \mSDF models geometry but is textureless. 
        To get $\dinofeat$ features, it applies the \dino model on the textured image $\teximg$ (applicable also for the mSDF, see next paragraph), to extract per-pixel geometric cues. 
        To form the final features, it concatenates \cite{NEURIPS2023_tale2features} per pixel the normalized $\diffusionfeat$ features with $\dinofeat$ ones, as $\cF = \{\alpha \diffusionfeat, (1-\alpha) \dinofeat\}$, where $\alpha$ is a steering weight.
        A detection mask, $\gtmask$, steers focus only on object pixels. 
        Below, the flattened features, $\maskedimgfeat$, are denoted as $\imgfeat$ for notational brevity.

\input{fig/05_template_query}

    \zheading{Shape (\mSDF) Features} 
    \label{subsub:shape_features}
        Recently, \difff~\cite{dutt2023diffusion} decorates 3D meshes with features extracted via \controlnet \cite{Zhang_2023_ICCV} and \dino\cite{oquab2024dinov2}. 
        \nameMethod follows this to obtain features for the textureless \mSDF and establish 2D-3D correspondences with image features in a zero-shot fashion. 
        This is a novel use of \difff for a long-standing problem. 
        Note that this does not require a known object-part connectivity~\cite{bogo2016simplify,wu2023magicpony}. 
        Note also that the DIT~\cite{dit} \mSDF model establishes dense correspondences across all morphed shapes within a class. 
        
        \nameMethod can perform the above in two different ways: 
        \qheading{(1)~``\nameMethod~feat@$\mathbf{T}$'' (\cref{fig:template_query}-left)} 
        It decorates a mesh extracted from the \mSDF template, $T$, only once per category, offline.
        Then, for every morphed \mSDF shape, it queries decoration features from the already decorated $T$. 
        \qheading{(2)~``\nameMethod~feat@$\mathbf{S_{init}}$'' (\cref{fig:template_query}-right)} 
        It decorates a mesh extracted from the initial \mSDF shape, $S_{init}$ (\cref{sub:shape_init}). 
        It does so once per image, as $S_{init}$ differs across images. 
        The above options trade efficiency for accuracy; the former is computationally cheaper, but the latter is more accurate.

        In any case, for decoration, \nameMethod first extracts a mesh \cite{shen2023flexicubes} from either $T$ or from $S_{init}$.
        Then, it samples $\nviews$ views on a unit sphere around it, and for each view $j \in \nviews$, it renders normal maps, $\prednormals^j$, and depth maps, $\preddepth^j$.  
        Then, it extracts per-pixel feature maps, $\shapefeat^{j}$, in the same way as for image features, discussed above. 
        Since the $P^j$ camera parameters are known, each view-specific feature map, $\shapefeat^{j}$, gets unprojected onto 3D mesh vertices. 
        Last, for each vertex, the unprojected features across views are aggregated to form the final feature, $\shapefeat \in \nR^{|S| \times \featdim}$, where $|S|$ is the number of vertices and $\featdim$ is the feature dimension.

    \zheading{Object-to-Image Alignment}
    \label{subsub:matching}
        Using the extracted image features, $\imgfeat$, and shape feature maps, $\shapefeat$, 
        \nameMethod establishes 2D-3D \imgtoshape correspondences, $\cC$, by finding in feature space the most similar vertex for each pixel:
        \vspace{-0.4 em}
        \begin{align}
        \label{eq:feat_matching}
            \cC = &
            \{
            \{i, s\} = \argmax_{s \in \initshape} \similaritymatrix_{i,s}, 
            \;\,
            \text{for all pixels $i$} 
            \}
            \text{.}
            \\
            \similaritymatrix_{i,s} = &
            \frac{\imgfeat^i \;\, \cdot \;\, \shapefeat^s}
            {\|\imgfeat^i\|_2  \;\, \;\,  \|\shapefeat^s\|_2}
            \text{,}
        \end{align}

        \noindent
        where 
        $\similaritymatrix$ is a cosine-similarity matrix, and 
        $\imgfeat^i$ and $\shapefeat^s$ are \mbox{$i$-th} pixel and \mbox{$s$-th} vertex features. 
        By exploiting the correspondences, $\cC$, \nameMethod implicitly finds the visible \mSDF points, as only these can be matched to pixels (see \cref{sec:supmat_matching}).

        Moreover, \nameMethod estimates intrinsic camera parameters, $K$, via the off-the-shelf \mbox{PerspectiveFields}~\cite{jin2022PerspectiveFields} model applied on image $\img$.
        Last, it uses the estimated correspondences, $\cC$, and intrinsics, $K$, to apply the \mbox{RANSAC}~\cite{fischler_bolles_1981} and \mbox{PnP}~\cite{pnp} algorithms for estimating the object pose, $\initR, \initT$. 
        This pose, along with the initial object shape, $\latinit$ (\cref{sub:shape_init}), initializes our fitting framework (\cref{sub:fitting}). 

%% file: fig/03_method.tex
\begin{figure*}
    \centering
        \includegraphics[width= 0.99 \textwidth]{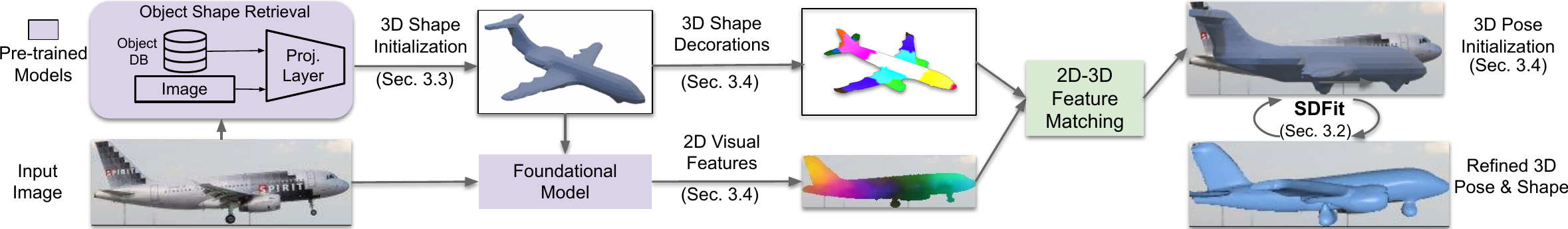}
        \vspace{-0.5 em}
        \caption{
            Our \nameMethod framework. 
            We represent 3D shape via a learned morphable signed-distance function (\mSDF) model \cite{dit} (\cref{sub:shape_prior}). 
            We first recover a likely initial shape from a database \cite{chang2015shapenet} via a \sota retrieval method~\cite{liu2023openshape} conditioned on the input image (\cref{sub:shape_init}). 
            Next, we extract features from both the target image and the initial shape via foundational models~\cite{oquab2024dinov2,Zhang_2023_ICCV} 
            to establish image-to-\mSDF 2D-to-3D correspondences and initialize pose (\cref{sub:pose_init}). 
            Last, we iteratively refine both shape and pose via render-and-compare (\cref{sub:fitting}). 
        }
    \label{fig:method}
    \vspace{-0.5 em}
\end{figure*}

%% file: fig/04_method_canonloss.tex
\begin{figure}
    \centering
        \includegraphics[width=1.0 \columnwidth]{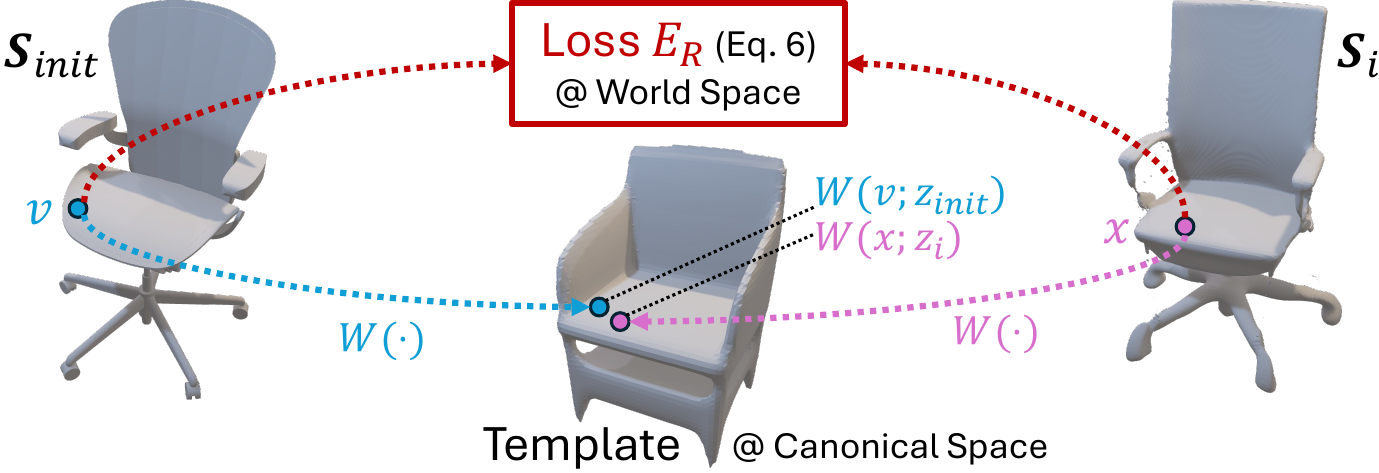}
        \vspace{-1.5 em}
        \caption{
            Loss $E_{R}$ (\cref{eq:regularization}). 
            Initial and hypothesis shapes, $S_{init}$ and $S_i$, are warped to a canonical space via \DIT's warper $W(\cdot)$. 
            For each warped vertex of $S_i$ we find the closest warped vertex of $S_{init}$ in canonical space, and compute MSE in world space.
        }
        \label{fig:method_canonloss}
\end{figure}

%% file: fig/05_template_query.tex
\definecolor{pptgreen}{RGB}{90,160,30}
\begin{figure}
    \centering
        \vspace{-0.25 em}
        \includegraphics[width=0.90 \columnwidth]{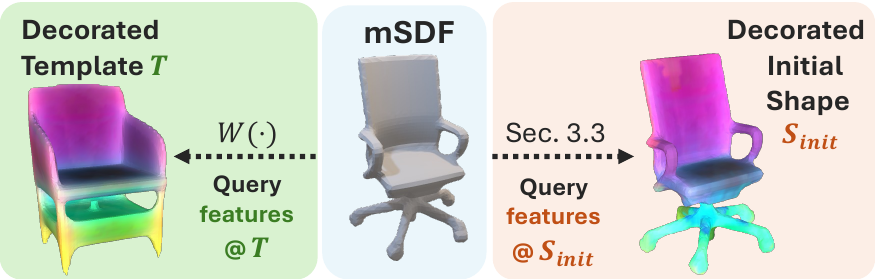}
        \vspace{-0.5 em}
        \caption{
            Features for \mSDF.
            Rather than decorating each \mSDF from scratch, we can query precomputed features via DIT’s correspondences and warper, $W(\cdot)$.
            We either pre-decorate the category template (\textcolor{ForestGreen}{\tt feat@$T$}) or the initial shape per image (\textcolor{orange}{\tt feat@$S_{init}$}).
    }
    \label{fig:template_query}
    \vspace{-0.2 em}
\end{figure}

%% file: sec/04_Results.tex
\section{Experiments}

\subsection{Implementation Details}
\label{sub:impl_details}

        \hspace{\parindent}
        \zheading{\mSDF}
        We use the \DIT~\cite{dit} model trained on \ShapeNet \cite{chang2015shapenet}. 
        But our approach is agnostic to the chosen \mSDF, that is, as richer \mSDFs get developed, \nameMethod also gets better.

        \zheading{Pose Initialization (\cref{sub:pose_init})}
        We establish image-to-shape 2D-3D correspondences by matching deep features. 
        However, these might be imperfect as this is still an open problem. 
        Therefore, we compute pose as follows. 
        First, we apply \mbox{RANSAC+PnP} on the established correspondences, and generate two hypotheses by mirroring pose around the vertical axis. 
        Then, we refine each hypothesis over 200 iterations and select the one with the lower $E_D$ from \cref{eq:objectiveTERMS}.

        \zheading{Normal, Depth \& Mask Maps}
        For the objective function of \cref{eq:objectiveFULL} we need an observed ``ground truth'' segmentation mask, $\gtmask$, normal map, $\gtnormals$, and depth map, $\gtdepth$, and respective maps rendered from the \mSDF, $\predmask$, $\preddepth$, and $\prednormals$.
        
        For $\predmask$, $\preddepth$, $\prednormals$, we extract a mesh via \flexicubes~\cite{shen2023flexicubes} with a grid size of $\griddim=\griddimnum$ and render with \mbox{Nvdiffrast}~\cite{Laine2020diffrast}. 
        
        We estimate $\gtnormals$ and $\gtdepth$ by applying the \mbox{OmniData}~\cite{kar20223d} model on the input image. 
        The masks $\gtmask$ can be provided by datasets (\eg, in \pixTD~\cite{pix3d}), while in the opposite case (\eg, for \pascalTD~\cite{pascal3d+}) we segment objects by applying the {\tt rembg} \cite{rembg} method (as in \zeroshape~\cite{huang2023zeroshape}).

        \zheading{Fitting (\cref{sub:fitting})}
        We optimize with Adam~\cite{adam}. 
        For the first $\niterpose$ iterations, we refine the initial pose, ($\initR, \initT$), and scale, $\initS$, keeping shape $\initshape$ fixed. 
        For the next $\nitershape$ iterations, we jointly optimize shape, scale and pose.

        \input{tab/01_02_occ_recon_and_runtime}

\subsection{Metrics}
\label{sub:eval_metrics}

    We use four complementary numeric metrics as follows.

    \zheading{Chamfer Distance (CD)} 
        CD quantifies the similarity of two 3D point clouds $X$ and $Y$ as the average (bidirectional) distance from each point in a cloud to the nearest point in the other one. 
        Then, with $|.|$ denoting cardinality: 
        {\small
        \begin{equation}
            CD= \frac{1}{|X|} \sum_{x \in X} \min_{y \in Y} \|x - y\|_2 + \frac{1}{|Y|} \sum_{y \in Y} \min_{x \in X} \|x - y\|_2
            \text{.}
        \end{equation}}

    \zheading{\FScore}
        Given a rejection threshold, $d$, the \FScore at distance $d$ (F@\textit{d}) is the harmonic mean of \textit{precision@d} and \textit{recall@d}, reflecting the proportion of the surface accurately reconstructed within the correctness threshold, $d$.

    \zheading{Intersection-over-Union (IoU)}
        IoU encodes the alignment of an estimated 3D shape with image pixels, by quantifying the alignment between a target mask (detected in image) and estimated mask (projected 3D shape onto 2D) as:
        {\small
        \begin{align}
            IoU = ({TP}) \;\ / \;\ ({TP+FP+FN}) \times 100
            \text{, \;\, where:}
        \end{align}}TP is true positives, FP false positives, FN false negatives.

    \zheading{CLIP Similarity}
        To assess how plausible 3D shapes look like for a given class (\eg, a ``chair''), we first compute the CLIP embedding of the class name and of a rendered 3D-shape image, and then their \mbox{CLIP} Similarity~\cite{clipscore}.

\subsection{Evaluation}
\label{sub:eval_tasks}

    We evaluate on \emph{shape reconstruction}, capturing only geometry, and \emph{image alignment}, capturing both shape and pose. 

    \zheading{Shape Reconstruction} 
    We evaluate on the standard \pixTD dataset \cite{pix3d}, which pairs real-world images with \groundtruth CAD models, using \zeroshape's \cite{huang2023zeroshape} test set. 
    Specifically, we compare our fitting-based \nameMethod method against 
    the regression-based \zeroshape~\cite{huang2023zeroshape} and \triposr~\cite{TripoSR2024}, and 
    the diffusion-based \sdfusion~\cite{cheng2023sdfusion} model. 
    For a more direct comparison, we also train a class-specific \zeroshape~\cite{huang2023zeroshape}, denoted as \classzeroshape. 

    \input{fig/06_qualitative}

    The results, presented in \cref{tab:pix3d_reconstruction}, show that \nameMethod performs on par with \zeroshape and \triposr in terms of the CD metric.
    However, we notice that \zeroshape often defaults to ``blobby'' shapes (see \cref{fig:comparison}). 
    We hypothesize that this might be because it is ``only'' feed-forward, so it cannot correct potential mistakes. 
    Moreover, its search space might be insufficiently constrained, so it often struggles to produce shapes resembling the depicted object class. 
    
    \input{tab/03_recon_pix3d}

    However, the CD metric cannot fully capture such artifacts, as it measures only geometric proximity to ground-truth shapes, and ignores semantics. 
    To capture semantics, we assess how well a recovered 3D shape aligns with the target class via \mbox{CLIP} similarity~\cite{clipscore,dreamhoi}. 
    To this end, we first compute the CLIP embedding for the class name. 
    Then, we render synthetic images of the fitted \mSDF from five canonical viewpoints, and compute their CLIP embedding. 
    Last, we compute the cosine-similarity score between the class embedding and the five \mSDF embeddings, taking the max score to account for poor views. 
    As shown in \cref{tab:pix3d_reconstruction}, \nameMethod yields shapes that better reflect the target class.

\zheading{Shape Reconstruction under Occlusion}
    We quantitatively evaluate robustness to occlusion by 
    (1)~rendering synthetic occluding patches on \pixTD images covering 40\% of the object's bounding box (\cref{fig:occluders}), and  
    (2)~using the \comic~\cite{chord} dataset of hand-object grasps.
    
    Results are shown in \cref{tab:occ_reconstruction}, and a sensitivity analysis in \cref{fig:occlusion_plot}, and \cref{sec:supmat_occlusion}.
    We see that \nameMethod clearly outperforms \zeroshape. 
    Note that \nameMethod has stable performance for increasingly stronger occlusions, while \zeroshape heavily degrades. 
    We think that this is because \nameMethod relies on geometric cues only from unoccluded regions, which remain intact, while \zeroshape relies on ``global'' appearance cues that are strongly influenced by occlusions. 
    Moreover, \nameMethod uses an explicit shape prior (\mSDF) and a feedback loop, while \zeroshape uses an implicit shape prior (baked into network weights) and does only feed-forward inference.
    
    To help baselines handle occlusions, we give them the privilege of inpainting~\cite{yu2023inpaint}.
    That is, we remove occluders, infill the missing object pixels~\cite{yu2023inpaint}, and apply the baseline on the new unoccluded image. 
    The privileged baselines have an improved performance, but \nameMethod clearly outperforms these. 
    This is because \nameMethod relies only on features from unoccluded regions, and leverages the \mSDF shape manifold of valid shapes to recover the occluded parts.

    We compare the best performers of \cref{tab:pix3d_reconstruction}, \ie, \nameMethod and \zeroshape~\cite{huang2023zeroshape}, in \cref{fig:comparison}. 
    \zeroshape struggles recovering self-occluded parts. 
    Instead, \nameMethod recovers these via the regularizer $E_R$ in \cref{eq:regLoss}. 
    This aligns with the previous paragraph, \ie, \nameMethod is more robust to self-occlusions or occlusions by third parties. 
    This is because \nameMethod exploits correspondences (see \cref{fig:template_query}) between $S_{init}$ and the running \mSDF hypothesis for supervising self-occluded regions. 

    \input{fig/07_occlusion_example}
    \input{fig/08_occlusion_plot}
    \input{fig/09_occlusion_qualitative}
    \input{fig/10_qualitative_sdfit}

    \zheading{Image Alignment} 
    We evaluate joint shape-and-pose estimation for inferring pixel-aligned 3D objects. 
    We use the \pascalTD~\cite{pascal3d+} dataset and specifically the test split of Pavllo~\etal~\cite{pavllo2021textured3dgan} for the {\tt car} and {\tt airplane} classes. 
    
    Since \sota methods \cite{huang2023zeroshape,cheng2023sdfusion} focus mostly on shape recovery, ignoring pose, we establish our own baselines, by extending these methods with our pose initialization and {\tt RnC} fitting as follows: 
    We first infer shape through \sota regression~\cite{huang2023zeroshape} or diffusion~\cite{cheng2023sdfusion} methods. 
    We then keep the estimated shape fixed, and optimize over pose and scale with our render-and-compare module ({\tt RnC}) (\cref{sub:fitting}).
    For fairness, we initialize the object pose and camera intrinsics of all baselines using \nameMethod's pose initialization (\cref{sub:pose_init}). 
    For \zeroshape, we initialize only the translation since it assumes that the world and camera frames are aligned.

    \Cref{tab:pascal3d_pix3d_iou} reports the 2D Intersection-over-Union (\%). 
    ``{\tt \nameMethod-feat@$S_{init}$}'' outperforms all baselines, while 
    ``{\tt \nameMethod-feat@$T$}'' is on par with ``{\tt \mbox{ZeroShape+RnC}}'' but outperforms others. 
    This shows that \nameMethod (whose {\tt RnC} module is used to extend baselines) and \zeroshape can be complementary. 
    Note that baselines refine only the pose through {\tt RnC}. 
    Instead, \nameMethod uniquely refines both pose and shape by morphing the \mSDF~-- this is a key advantage.

    Note that {\tt \nameMethod-feat@$T$} trades speed for accuracy; 
    it is faster but less accurate than {\tt \nameMethod-feat@$S_{init}$}, as the topology of $S_{init}$ matches the image better than $T$.
    For further ablations of our modules, see \cref{sec:ablation-modules}
    in \supmat

    \zheading{Qualitative Results} 
    We show extensive results of our {\tt \nameMethod-feat@$S_{init}$} for \inthewild \pascalTD~\cite{pascal3d+} and \pixTD~\cite{pix3d} images in \cref{fig:results_qualitative}. 
    Despite the diverse shapes, appearances, and challenging imaging conditions (\eg, poor lighting, uncommon poses) in real-world images, \nameMethod recovers plausible, pixel-aligned 3D shapes, showing promising generalization.
    Unlike purely data-driven methods, \nameMethod does not need retraining for unseen images. 
    
    Moreover, in \cref{fig:occlusion_qualitative} we show reconstructions of \nameMethod \emph{under occlusion} on \pixTD \cite{pix3d} and \comic~\cite{chord}, as well as on \mbox{COCO}~\cite{mscoco} images that are taken in the wild. 
    \nameMethod's reconstructions look robust to strong occlusions, reflecting the findings of \cref{tab:occ_reconstruction,fig:occlusion_plot}, and \supmat \cref{sec:supmat_occlusion}.

    \input{tab/04_alignment}

\pagebreak
\zheading{Runtime}
    Our \nameMethod method fully converges in $\sim$\avgtime, often obtaining a satisfactory result within the first 45--60 sec, using an \mbox{Nvidia} \mbox{4080} GPU, with an additional 20 sec for image feature extraction \cite{Zhang_2023_ICCV, oquab2024dinov2}, and 10 sec for shape decoration. 
    For the latter, see \cref{tab:supmat_decorating}, where the top-most row corresponds to \difff~\cite{dutt2023diffusion}; 
    heavily reducing the number of views and diffusion steps does not harm accuracy, while reducing runtime 60x \wrt \difff. 
    We hypothesize that \difff's many views introduce redundant information that causes noise accumulation during feature aggregation.

%% file: tab/01_02_occ_recon_and_runtime.tex
\definecolor{paleyellow}{rgb}{1.0, 1.0, 0.8}
\begin{table*}
    \centering
    \vspace{-0.5 em}
    \begin{minipage}{0.68\textwidth}
        \centering
        \resizebox{\textwidth}{!}{
            \begin{tabular}{lccccccccccc} 
                \toprule
                {Metric: CD $\downarrow$} & & \multicolumn{3}{c}{{Pix3D}} & \multicolumn{7}{c}{{COMIC}} \\ 
                \cmidrule(lr){3-5} \cmidrule(lr){6-12}
                Method & Inpaint & {\tt Chair} & {\tt Sofa} & {Mean} & {\tt Box} & {\tt Bottle} & {\tt Camera} & {\tt Knife} & {\tt Mug} & {\tt Spray} & {Mean} \\ 
                \midrule
                \ZeroShape~\cite{huang2023zeroshape} & \xmark & 7.08 & 5.71 & 6.40 & 9.32 & 5.37 & 9.25 & 3.24 & 7.31 & 4.67 & 6.53 \\
                \classzeroshape~\cite{huang2023zeroshape} & \xmark  & 5.94 & 5.14 & 5.54 & 7.14 & 4.90 & 6.40 & 1.98 & 4.44 & 4.90 & 4.96 \\
                \triposr~\cite{TripoSR2024} & \xmark & 6.73 & 7.45 & 7.09 & 10.7 & 6.48 & 9.66 & 3.28 & 6.23 & 4.81 & 6.86 \\
                \midrule
                \ZeroShape~\cite{huang2023zeroshape} & \cmark & 5.64 & 4.81 & 5.23 & 7.91 & 4.38 & 8.42 & 2.77 & 6.52 & 3.83 & 5.64 \\
                \classzeroshape~\cite{huang2023zeroshape} & \cmark & 5.71 & 4.83 & 5.27 & 6.96 & 4.38 & 6.05 & 1.55 & 3.92 & 6.54 & 4.90 \\
                \triposr~\cite{TripoSR2024} & \cmark & 6.03 & 6.95 & 6.49 & 6.39 & 4.25 & 7.06 & 2.19 & 4.91 & 4.60 & 4.90 \\
                \midrule 
                \nameMethod-{\tt $\text{feat@}S_{init}$} & \xmark & \textbf{4.08} & \textbf{3.40} & \textbf{3.74} & \textbf{1.21} & \textbf{4.04} & \textbf{6.52} & \textbf{1.11} & \textbf{2.94} & \textbf{3.66} & \textbf{3.25} \\ 
                \bottomrule
            \end{tabular}
        }
        \vspace{-0.7 em}
        \caption{
            \textbf{Shape reconstruction} evaluation \textbf{under occlusion} on \pixTD~\cite{pix3d} (synthetic-patch occlusions) and \comic~\cite{chord} (hand-object grasp occlusions).
            For \pixTD we occlude 40\% of the object bounding box (see \cref{sec:supmat_occlusion} in \supmat).
            We report the Chamfer Distance (CD).
        }
        \label{tab:occ_reconstruction}
    \end{minipage}
    \vspace{-0.5 em}
    \hfill
    \begin{minipage}{0.3\textwidth}
        \centering
        \resizebox{\textwidth}{!}{
            \begin{tabular}{ccc|c}
                \toprule
                {\# Views} & {Diff. steps} & {Runtime (sec)} & CD $\downarrow$ \\
                \midrule
                100 & 100 & 600 & \textbf{3.53} \\ 
                16  & 10  & 27  & 3.67 \\
                8   & 30  & 30  & 3.69 \\ 
                \rowcolor{paleyellow}
                8   & 10  & 10  & 3.58 \\ 
                \bottomrule
            \end{tabular}
        }
        \vspace{-0.5 em}
        \caption{
            Runtime analysis for shape decoration. 
            We assess the impact of the number of views, diffusion steps, and runtime for  \texttt{Chair} and \texttt{Sofa} in \pixTD~\cite{pix3d}. 
        }
        \label{tab:supmat_decorating}
    \end{minipage}
    \vspace{-0.7 em}
\end{table*}

%% file: fig/06_qualitative.tex
\begin{figure}
    \vspace{-0.5 em}
    \centering
        \begin{overpic}[trim=000mm 002mm 000mm 005mm, clip=true, width=0.99\columnwidth,unit=1bp,tics=10]
        {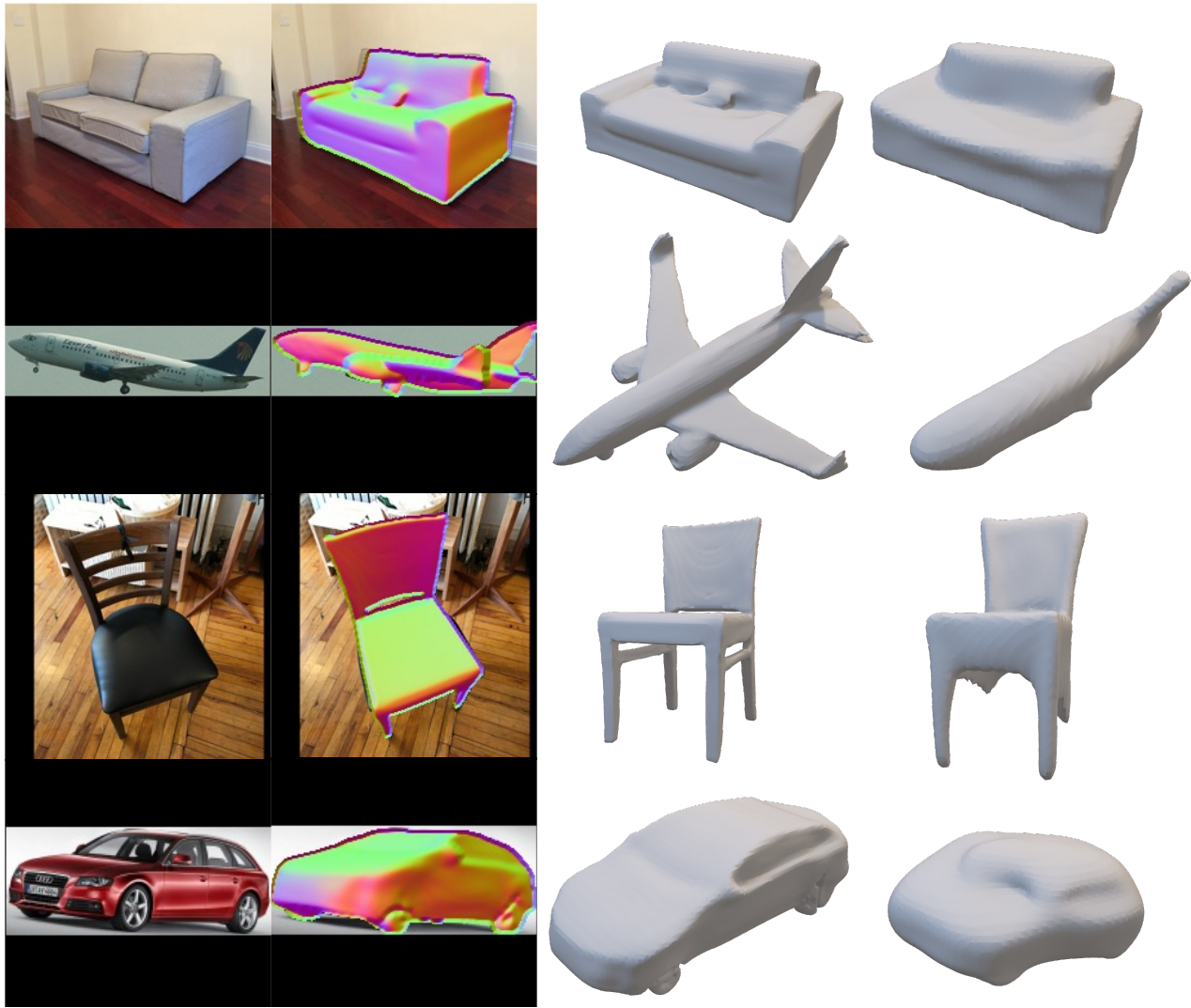}
     	\put(5,-5){\colorbox{white}{\parbox{0.10\textwidth}{\small Image}}}
            \put(30,-5){\colorbox{white}{\parbox{0.10\textwidth}{\small \hspace{-0.9 em} \nameMethod}}}
            \put(52,-5){\colorbox{white}{\parbox{0.10\textwidth}{\small \nameMethod}}}
            \put(75,-5){\colorbox{white}{\parbox{0.10\textwidth}{\small \hspace{-0.9 em} \zeroshape~\cite{huang2023zeroshape}}}}
        \end{overpic}
        \vspace{+0.7 em}
        \caption{
            Shape recovery for \nameMethod~({\tt $\text{feat@}S_{init}$}) and \zeroshape \cite{huang2023zeroshape}. 
            \nameMethod jointly fits pose and shape to the image, helping pixel alignment. 
            It also excels at recovering occluded parts via the \mSDF's learned shape prior. 
            Overlays show the \mSDF's normals.
        }
        \vspace{-0.5 em}
    \label{fig:comparison}
\end{figure}

%% file: tab/03_recon_pix3d.tex
\begin{table}
    \centering
    \resizebox{0.42 \textwidth}{!}{
        \begin{tabular}{llccccc} 
            \toprule
            {} &  Type & CD $\downarrow$ & F@1 $\uparrow$ & F@2 $\uparrow$ & CLIP$\cdot10^2\uparrow$ \\ 
            \midrule
            \red{\SDFusion}~\cite{cheng2023sdfusion} & Diff. & 3.95 & 0.16 & 0.38 & 29.87  \\
            \ZeroShape~\cite{huang2023zeroshape} & Reg. & 3.44 & 0.23 & \textbf{0.47} & 29.33 \\ 
            \classzeroshape~\cite{huang2023zeroshape} & Reg. & 4.55 & 0.21 & 0.46 & 29.40 \\ 
            \triposr~\cite{TripoSR2024} & Reg. & \textbf{3.41} & 0.13 & 0.32 & 29.87 \\ 
            \midrule
            \nameMethod-{\tt $\text{feat@}S_{init}$} & Opt. & 3.53 & \textbf{0.25} & 0.46 & \textbf{30.53} \\ 
            \bottomrule
        \end{tabular}
    }
    \vspace{-0.5 em}
    \caption{
        \textbf{Shape reconstruction} evaluation on \pixTD~\cite{pix3d}. 
        We report the mean Chamfer Distance (CD), F-Score at two thresholds (F@1 and F@2), and CLIP similarity across the {\tt Chair} and {\tt Sofa} categories; each value is the average over these classes. 
    }
    \label{tab:pix3d_reconstruction}
\end{table}

%% file: fig/07_occlusion_example.tex
\begin{figure}
    \centering
    \includegraphics[width=0.99 \columnwidth]{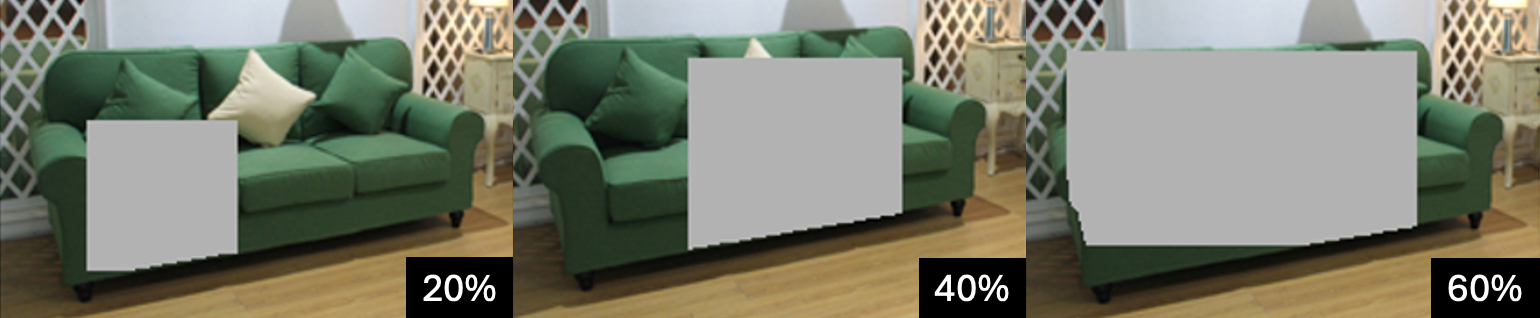}
    \vspace{-0.5 em}
    \caption{
        Examples of synthetic occluders of varying size. 
        The labels denote the percentage of object (bounding-box) occlusion. 
    }
    \label{fig:occluders}
\end{figure}

%% file: fig/08_occlusion_plot.tex
\begin{figure}
    \centering
        \vspace{-0.5 em}
        \includegraphics[width=0.9 \columnwidth]{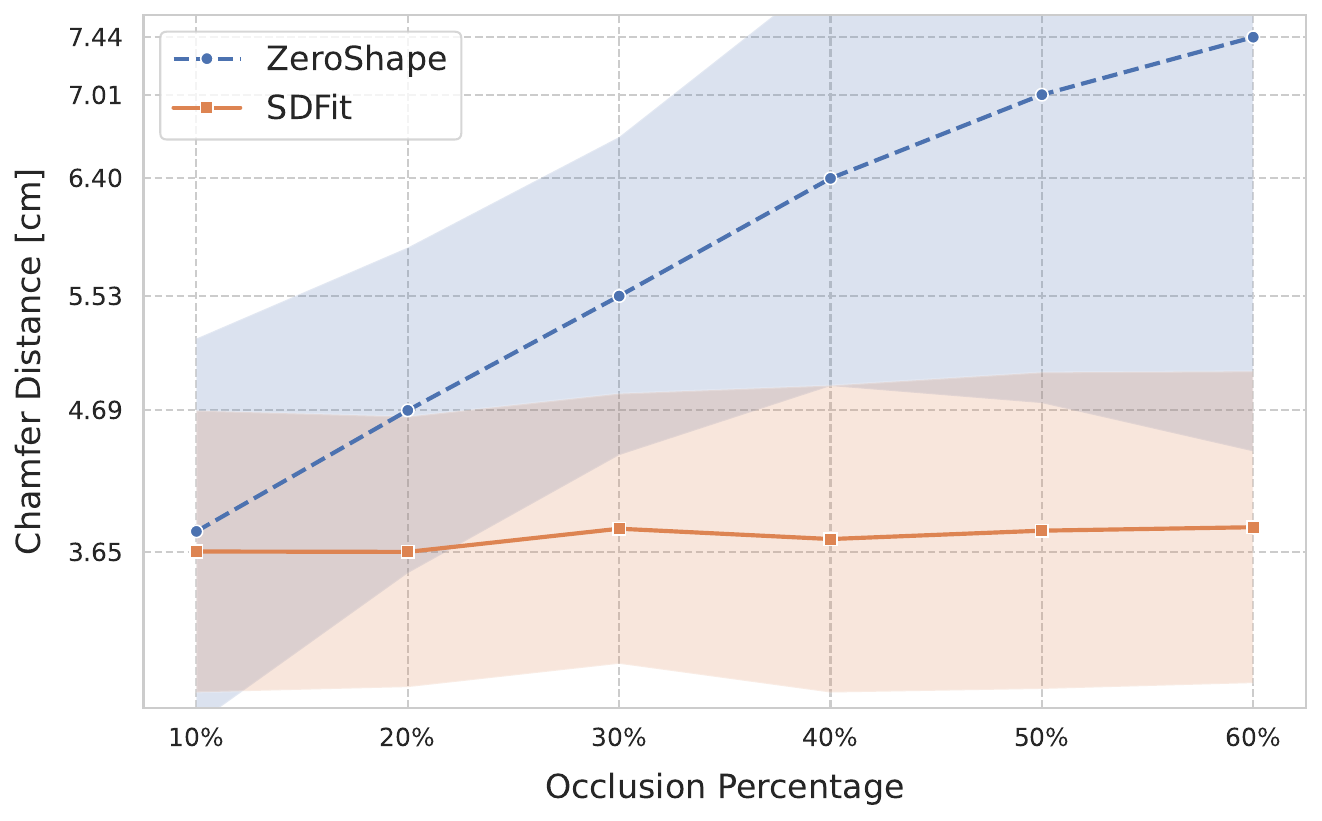}
        \vspace{-1.0 em}
        \caption{
            Occlusion sensitivity analysis. 
            We evaluate shape reconstruction (Y-axis) on the \pixTD~\cite{pix3d} test set with a varying degree of occlusion (X-axis). 
            \nameMethod outperforms \zeroshape in both mean and standard deviation (lower is better) and remains stable under increasing occlusion, while \zeroshape heavily degrades. 
    }
    \label{fig:occlusion_plot}
\end{figure}

%% file: fig/09_occlusion_qualitative.tex
\begin{figure}
    \centering
        \vspace{-0.5 em}
        \begin{overpic}[width=0.75 \columnwidth,unit=1bp,tics=10,grid=False]{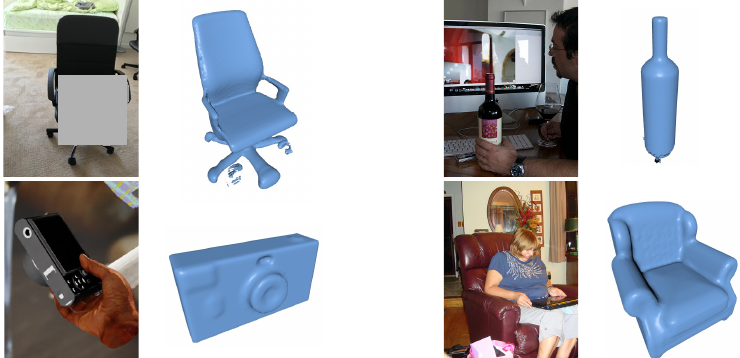}
            \put(-6.5,44){\parbox{0.10\textwidth}{\small (a)}}
            \put(-6.5,20){\parbox{0.10\textwidth}{\small (b)}}
            \put(53.5,44){\parbox{0.10\textwidth}{\small (c)}}
            \put(53.5,20){\parbox{0.10\textwidth}{\small (d)}}
        \end{overpic}
        \vspace{-0.5 em}
        \caption{
            Reconstructions of \nameMethod-{\tt $\text{feat@}S_{init}$} on images of three datasets: 
            (a) \pixTD~\cite{pix3d} with synthetic occluding patches, used in \cref{fig:occlusion_plot} and \cref{tab:occ_reconstruction}, 
            (b) COMIC~\cite{chord}, and 
            (c, d) COCO~\cite{mscoco}.
        }
        \vspace{-1 em}
    \label{fig:occlusion_qualitative}
\end{figure}

%% file: fig/10_qualitative_sdfit.tex
\begin{figure*}
    \centering
        \vspace{-0.5 em}
        \begin{overpic}[width=0.92 \textwidth,unit=1bp,tics=10,grid=False]
        {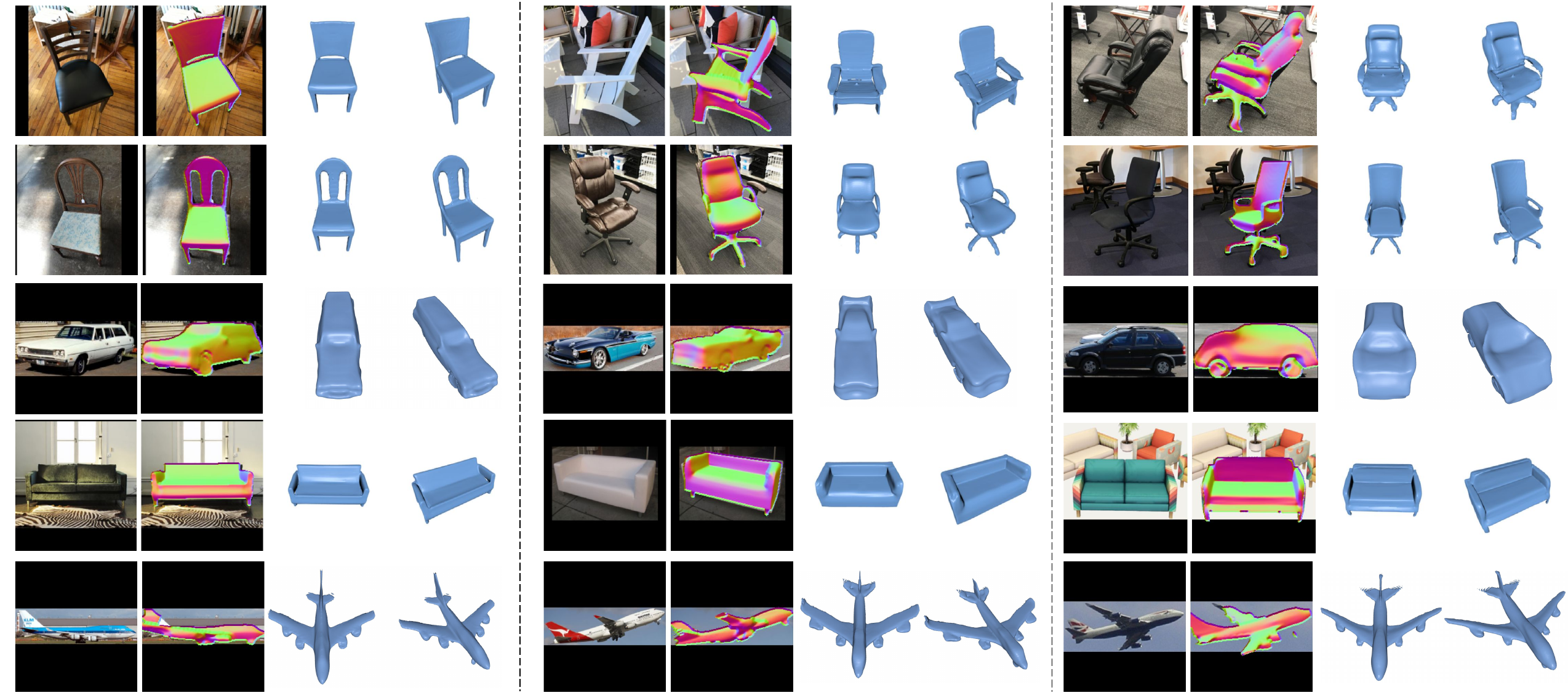}
     	\put(3,-1.5){\parbox{0.10\textwidth}{\small Image}}
            \put(10,-1.5){\parbox{0.10\textwidth}{\small Overlay}}
            \put(22,-1.5){\colorbox{white}{\parbox{0.10\textwidth}{\small \nameMethod}}}
            \put(36,-1.5){\parbox{0.10\textwidth}{\small Image}}
            \put(44,-1.5){\parbox{0.10\textwidth}{\small Overlay}}
            \put(57,-1.5){\parbox{0.10\textwidth}{\small \nameMethod}}
            \put(69,-1.5){\parbox{0.10\textwidth}{\small Image}}
            \put(76.5,-1.5){\parbox{0.10\textwidth}{\small Overlay}}
            \put(90,-1.5){\parbox{0.10\textwidth}{\small \nameMethod}}
        \end{overpic}
        \vspace{+0.5 em}
        \caption{
            Qualitative results for \nameMethod({\tt $\text{feat@}S_{init}$}) on images of the \pixTD~\cite{pix3d} and \pascalTD~\cite{pascal3d+} datasets. 
            We show the estimated 3D shape (left to right) in camera- (as normal map), front- and side-view. 
            \faSearch~\textbf{Zoom in} to see details.
        }
    \label{fig:results_qualitative}
    \vspace{-0.6 em}
\end{figure*}

%% file: tab/04_alignment.tex
\begin{table}
        \centering
        \vspace{-0.5 em}
        \resizebox{0.46\textwidth}{!}{
            \begin{tabular}{lccccc} 
                \toprule
                Metric: 2D IoU (\%)~$\uparrow$ & \multicolumn{2}{c}{Pascal3D+} & \multicolumn{2}{c}{Pix3D} & Mean \\ 
                \cmidrule(lr){2-3} \cmidrule(lr){4-5}
                & {\tt Plane} 
                & {\tt Car} 
                & {\tt Chair} 
                & {\tt Sofa} 
                & IoU \\ 
                \midrule
                \SDFusion~\cite{cheng2023sdfusion} {\tt + RnC} & \NA & \NA & 59.5 & \NA & \NA \\
                \OpenShape~\cite{liu2023openshape} {\tt + RnC} & 48.5 & 79.5 & 54.3 & 85.2 & 66.8 \\
                \ZeroShape~\cite{huang2023zeroshape} {\tt + RnC} & 76.4 & 89.1 & 61.3 & 85.9 & 78.1 
                \\
                \midrule
                \nameMethod~{\tt $\text{feat@}T$} & 77.6 & 86.9 & 61.1 & 89.7 & 78.8 
                \\ 
                \nameMethod~{\tt $\text{feat@}S_{init}$} & \textbf{81.5} & \textbf{91.6} & \textbf{75.9} & \textbf{92.7} & \textbf{85.4} 
                \\ 
                \bottomrule
            \end{tabular}
        }
        \vspace{-0.5 em}
        \caption{
            Image-alignment performance on the \pascalTD~\cite{pascal3d+} and Pix3D~\cite{pix3d} datasets. 
            The shape predictions of competing methods are aligned to the image in a render-and-compare ({\tt RnC}) fashion similarly to our \nameMethod. 
            We report the per-category IoU metric, as well as the Mean IoU across all categories. 
        }
        \label{tab:pascal3d_pix3d_iou}
        \vspace{-0.5 em}
    \end{table}

%% file: sec/05_Conclusion.tex
\section{Conclusion}

We develop \nameMethod, a novel method for fitting an explicit morphable 3D shape prior to single images. 
This uniquely refines both shape and pose using an explicit feedback loop. 
This achieves much better pixel alignment than \sota methods, and is exceptionally robust to occlusions. 
We believe that this is interesting for the broader 3D community and will inspire work that combines the best of our work and learning-based models. 
To this end, our code is available. 

%% file: sec/06_Ack_Discl.tex
\section{Acknowledgments \& Disclosure}

\qheading{Acknowledgements} 
We thank Bo\v{z}idar Anti\'{c}, Yuliang Xiu and Muhammed Kocabas for useful insights. 
We acknowledge \mbox{EuroHPC} JU for awarding the project ID \mbox{EHPC-AI-2024A06-077} access to \mbox{Leonardo} \mbox{BOOSTER}. 
This work also used the Dutch national e-infrastructure with the support of the SURF Cooperative using grant no. EINF-7589.
This work is partly supported by the ERC Starting Grant (project \mbox{STRIPES}, \mbox{101165317}, PI: D. Tzionas). 

\qheading{Disclosure} 
D. Tzionas has received a research gift from Google, and from the NVIDIA Academic Grant Program.

%% file: sec/XX_SupMat.tex
\clearpage
\setcounter{page}{1}
\maketitlesupplementary

\renewcommand{\thesection}{S.\arabic{section}}
\renewcommand{\thefigure}{S.\arabic{figure}}
\renewcommand{\thetable}{S.\arabic{table}}
\renewcommand{\theequation}{S.\arabic{equation}}
\setcounter{section}{0}
\setcounter{figure}{0}
\setcounter{table}{0}
\setcounter{equation}{0}

\input{fig/S1_supmat_feat_matching}

\section{2D-3D Pixel-Vertex Matching}
\label{sec:supmat_matching}
    The task of single-image 3D pose and shape estimation presents significant challenges due to depth ambiguities, and (self-)occlusions. 
    To address these issues, we propose a zero-shot pose initialization technique leveraging deep foundational features~\cite{oquab2024dinov2,Zhang_2023_ICCV}, inspired by image-to-image \mbox{(2D-2D)} matching methods~\cite{NEURIPS2023_tale2features}.

    Starting from a shape initialization obtained via our procedure (see \cref{sub:shape_init}), the goal is to establish \mbox{2D-to-3D} correspondences by matching 2D pixels to 3D points of the \mSDF. 
    Using a pre-trained \controlnet~\cite{Zhang_2023_ICCV} and \dino~\cite{oquab2024dinov2} model we extract feature descriptors for the 2D image, $\imgfeat$, and 3D shape, $\shapefeat$, as detailed in \cref{sub:pose_init}. 
    These descriptors are matched via cosine similarity (\cref{eq:feat_matching}) to obtain a set of \mbox{2D-to-3D} \imgtoshape correspondences.
    
    By leveraging the semantic and geometric cues encoded in the features of \controlnet and \dino~\cite{banani2024probing}, our approach implicitly identifies the visible 3D vertices from 2D pixels. 
    Examples of these matches are shown in \cref{fig:feat_matching}, where these are color-coded via the PCA of $\imgfeat$.

\section{Occlusion Sensitivity}
\label{sec:supmat_occlusion}
    As discussed in \cref{sub:eval_tasks} in paragraph ``Shape Reconstruction under Occlusion,'' we evaluate robustness under occlusion by performing a sensitivity analysis against \zeroshape~\cite{huang2023zeroshape}. 
    Specifically, we augment \pixTD~\cite{pix3d} test images by randomly rendering rectangle occluders covering varying percentages (from $10\%$ to $60\%$) of the object bounding box; see examples in \cref{fig:occluders}. 

    In the main paper we report the results in a plot (\cref{fig:occlusion_plot}). 
    Here we report the numerical values that correspond to this plot in terms of the Chamfer Distance metric -- see \cref{tab:supmat_occlusion}. 
    
    \nameMethod consistently outperforms \zeroshape for all occlusion levels (both in terms of mean error and st.~dev.), preserving object coherence even with substantial occlusion. 
    Notably, \zeroshape struggles even with minor occlusions (10\%-20\%), emphasizing \nameMethod's practical advantage.

\pagebreak

\input{tab/S1_supmat_occlusion_vertical}

\section{Ablation of \nameMethod Modules}
\label{sec:ablation-modules}
    We replace our shape- and pose-estimation modules with GT information, and report the 2D IoU (\%) on the \pixTD dataset similar to \cref{tab:pascal3d_pix3d_iou}. 
    
    We compare three methods: 
    (1) {\tt \nameMethod} that refines both shape and pose and achieves an IoU of 84.3\%, 
    (2) {\tt \nameMethod-poseGT} that refines only shape and achieves 85.6\%, and 
    (3) {\tt \nameMethod-shapeGT} that refines only pose and achieves 79.4\%. 
    
    This shows that \nameMethod performs on par with privileged baselines. 
    All variants clearly outperform {\tt \mbox{ZeroShape+RnC}} that achieves 73.3\%.

\section{Discussion \& Future Work}

    We leverage foundational features for pose initialization. 
    As common in existing work~\cite{Zhang_2024_CVPR}, sometimes there might be potential left-right ambiguities that we tackle by evaluating two vertically mirrored candidates. 
    Future work will explore more involved approaches, \eg, via learned regression or by directly lifting 2D features into 3D via metric depth~\cite{depthpro}.
    
    Moreover, sometimes fine details may be missed, as in other neural-field-based methods~\cite{cheng2023sdfusion,huang2023zeroshape}, due to the fixed resolution grid used for mesh extraction. 
    Future work will look into dynamically adapting resolution, or enhancing the \mSDF expressiveness with a more ``flexible'' latent space.

%% file: fig/S1_supmat_feat_matching.tex
\definecolor{pptgreen}{RGB}{90,160,30}
\begin{figure}
    \centering
    \includegraphics[trim=000mm 000mm 01.5mm 000mm, clip=true, width=0.99 \columnwidth]{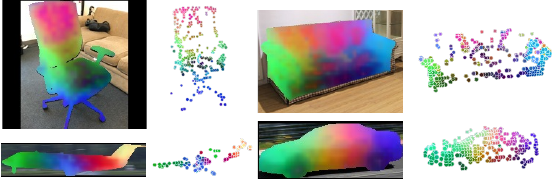}
    \vspace{-0.5 em}
    \caption{
        Feature matching examples. 
        For each pair -- 
        Left: PCA color-coded image features ($\imgfeat$). 
        Right: Corresponding \mSDF 3D points ($\shapefeat$), colored according to matched image pixels.
    }
    \vspace{-0.5 em}
    \label{fig:feat_matching}
\end{figure}

%% file: tab/S1_supmat_occlusion_vertical.tex
\begin{table}
    \centering
    \resizebox{0.80 \columnwidth}{!}{
    \begin{tabular}{lcc}
        \toprule
        Occlusion & \multicolumn{2}{c}{Pix3D (mean CD@XX) $\downarrow$} \\
        \cmidrule(lr){2-3}
        (\%) & \ZeroShape~\cite{huang2023zeroshape} & \nameMethod (Ours) \\
        \midrule
        0\%  & \textbf{3.44}$\pm$1.45 & 3.53$\pm$\textbf{0.82} \\
        10\% & 3.80$\pm$1.42 & \textbf{3.66}$\pm$\textbf{1.03} \\
        20\% & 4.69$\pm$1.20 & \textbf{3.65}$\pm$\textbf{0.99} \\
        30\% & 5.53$\pm$1.17 & \textbf{3.82}$\pm$\textbf{1.00} \\
        40\% & 6.40$\pm$1.53 & \textbf{3.74}$\pm$\textbf{1.13} \\
        50\% & 6.76$\pm$1.83 & \textbf{3.74}$\pm$\textbf{1.23} \\
        60\% & 7.45$\pm$2.48 & \textbf{3.83}$\pm$\textbf{1.15} \\
        \bottomrule
    \end{tabular}
    }
    \vspace{-0.5 em}
    \caption{
        Sensitivity analysis on occlusion. 
        We evaluate reconstruction accuracy under varying occlusion levels on the \pixTD~\cite{pix3d} test set, reporting the mean and standard deviation of Chamfer Distance (CD). 
        We also show the case with 0\% occlusion (result from \cref{tab:pix3d_reconstruction}) as reference. 
        Note that the occlusion percentage is computed on bounding boxes (that might be non-tight for the depicted object), so 60\% corresponds to excessively strong occlusions; see examples in \cref{fig:occluders}. 
        \nameMethod consistently outperforms \zeroshape (ZS), demonstrating greater stability and robustness as occlusion increases, whereas \zeroshape heavily deteriorates.
    }
    \label{tab:supmat_occlusion}
    \vspace{+1.2 em}
\end{table}